# Improving Interactive Diagnostic Ability of a Large Language Model Agent Through Clinical Experience Learning

[1]Zhoujian Sun, [2]Ziyi Liu, [1]Cheng Luo, [3]Jiebin Chu, [4]Zhengxing Huang

1 Research Center of Computational Life Science, Zhejiang Lab
2 TransTek Medical
3 Sir Run Run Shaw Hospital, School of Medicine, Zhejiang University
4 College of Computer Science and Technology, Zhejiang University

Corresponding author: Zhengxing Huang
zhengxinghuang@zju.edu.cn

# Abstract

Recent advances in large language models (LLMs) have shown promising results in medical diagnosis, with some studies indicating superior performance compared to human physicians in specific scenarios. However, the diagnostic capabilities of LLMs are often overestimated, as their performance significantly deteriorates in interactive diagnostic settings that require active information gathering. This study investigates the underlying mechanisms behind the performance degradation phenomenon and proposes a solution. We identified that the primary deficiency of LLMs lies in the initial diagnosis phase, particularly in information-gathering efficiency and initial diagnosis formation, rather than in the subsequent differential diagnosis phase. To address this limitation, we developed a plug-and-play method enhanced (PPME) LLM agent, leveraging over 3.5 million electronic medical records from Chinese and American healthcare facilities. Our approach integrates specialized models for initial disease diagnosis and inquiry into the history of the present illness, trained through supervised and reinforcement learning techniques. The experimental results indicate that the PPME LLM achieved over 30% improvement compared to baselines. The final diagnostic accuracy of the PPME LLM in interactive diagnostic scenarios approached levels comparable to those achieved using complete clinical data. These findings suggest a promising potential for developing autonomous diagnostic systems, although further validation studies are needed.

# 1 Introduction

Developing an autonomous disease diagnosis agent has been a key objective since the emergence of medical artificial intelligence (AI) [1]. Recent advances in large language models (LLMs) have opened up unprecedented possibilities for AI-assisted diagnosis. Current LLMs perform well on various assessments of medical abilities, successfully passing medical licensing examinations in different countries [2,3]. Moreover, several recently published controlled trial studies claim that leading LLMs can outperform human physicians in specific diagnostic scenarios [4-7].

Despite this progress, it is acknowledged that the performance of LLMs in diagnostic tasks may be systematically overestimated. Most current studies evaluate LLMs through single-turn question-answering tasks, where diagnostic information is readily available [4-8]. This setting sharply contrasts with real-world medical practice, where diagnosis is an iterative process requiring active information gathering through multi-turn interactions [9]. When tasked with proactively collecting clinical information from scratch, LLMs experience a significant decrease in diagnostic accuracy, with some studies reporting performance drops exceeding 50% [10-14]. This limitation indicates that current LLMs cannot handle diagnostic tasks independently. Previously, the performance decline of LLMs in interactive scenarios has been vaguely described as "LLMs lack planning capabilities." To our knowledge, the specific mechanisms underlying this limitation remain insufficiently investigated, and potential solutions have yet to be systematically explored [10-14].

This study presents an analysis of the performance degradation phenomenon and proposes a remedial approach. Our investigation was based on a widely accepted conceptual model that divides the physician diagnostic process into two distinct phases [15,16]. This model defines the first phase as experience-based patient history collection and high-risk disease identification, while the second phase involves applying clinical knowledge for the differential diagnosis of high-risk diseases identified in the first phase. Through evaluating LLMs' performance in both phases, we found that the primary deficiency in their planning capability lies in the first phase. Specifically, we observed that LLMs struggle with gathering information efficiently and have difficulty establishing accurate initial diagnoses based on insufficient medical history. However, we noted that LLMs can identify the actual disease through multiple rounds of differential diagnosis interactions when provided with a predefined list of high-risk diseases. LLMs' performance in this context can even significantly surpass what they achieve when directly reading complete patient information. Therefore, we propose that the statement "LLMs lack planning capability" can be more accurately expressed as: LLMs are not skilled at inquiring about medical history and inferring high-risk diseases. This limitation consequently undermines LLMs' ability to seek out essential diagnostic information, especially regarding laboratory tests and radiological findings, leading to poor final diagnostic performance. LLMs appear to show acceptable planning ability in differential diagnosis when presented with a predetermined list of high-risk diseases. Therefore, the primary challenge in enhancing the interactive diagnostic capabilities of LLMs is improving their performance in the first phase.

We argue that the primary factor contributing to the unsatisfactory performance of LLMs in the first phase arises from insufficient clinical experience. As diagnostic information remains incomplete during this phase, physicians rely more on hands-on clinical experience than on literature-derived knowledge or logical reasoning when gathering medical history and establishing initial diagnoses [15,16]. Ideally, LLMs would gain experience through extensive exposure to high-quality physician-patient dialogue data. However, due to various limitations, authentic medical consultation dialogues are rarely documented or used for training purposes [17]. This lack of expert demonstrations leads to LLMs lacking experiential intuition. Related studies addressing this issue have primarily focused on fine-tuning LLMs using either online medical consultation data or synthetic dialogues [18,19]. These methods encounter three main limitations: restricted dataset size, inadequate quality, and inherent statistical biases [20]. Given the significant challenges in acquiring comprehensive, high-quality physician-patient dialogue datasets, we explore a method to enable experiential learning for efficient initial diagnosis establishment.

Machine learning (ML) has achieved considerable success in fields like Go and video games, illustrating that ML algorithms can develop advanced decision-making skills through extensive trials [21-23]. This accomplishment suggests that we may not necessarily need to learn interactive policies from actual dialogue data. ML algorithms could independently acquire inquiry and initial disease diagnostic capabilities by generating diverse and realistic simulated patients. Electronic medical records (EMRs) are valuable for developing such simulation patients. Firstly, many hospitals have amassed millions of EMRs over time, providing a cost-effective and extensive data foundation. Secondly, EMRs are meticulously documented by healthcare professionals, containing crucial information necessary for disease diagnosis. Thirdly, nearly all patients seeking medical care produce an EMR, making these records a comprehensive and unbiased representation of real-world patient populations. These attributes effectively address the limitations of online consultations and synthetic datasets. Therefore, we aim to enhance the interactive diagnostic capabilities of LLM agents by leveraging EMR data.

While EMRs hold the potential to create simulated patients, training interactive policies through ML continues to be challenging. Existing reinforcement learning (RL) methods necessitate extensive iterations to optimize LLM parameters and develop new policies, with the cost of LLM fine-tuning remaining prohibitively high [24]. To overcome this limitation, we propose a plug-and-play method (Figure 1). In the training phase, our interdisciplinary team conducted a systematic review of standard medical textbooks to create a comprehensive checklist for the history of present illness (HPI) for common conditions, along with a standardized list of questions for HPI data collection. We compiled datasets that include over 3.5 million EMRs from both Chinese and American healthcare facilities. We established an LLM-based preprocessing pipeline to efficiently structure these HPI records and extract additional historical information, including demographics, past medical history (PMH), social history (SH), and family history (FH), among others, from EMRs. Subsequently, we developed two specialized models: a disease-diagnosis model trained using supervised learning (SL) techniques and an HPI inquiry model trained with an RL algorithm. We trained the inquiry model utilizing structured HPI and questioning lists. These components were integrated as plugins into the LLM framework, enhancing its initial disease diagnostic capabilities interactively.

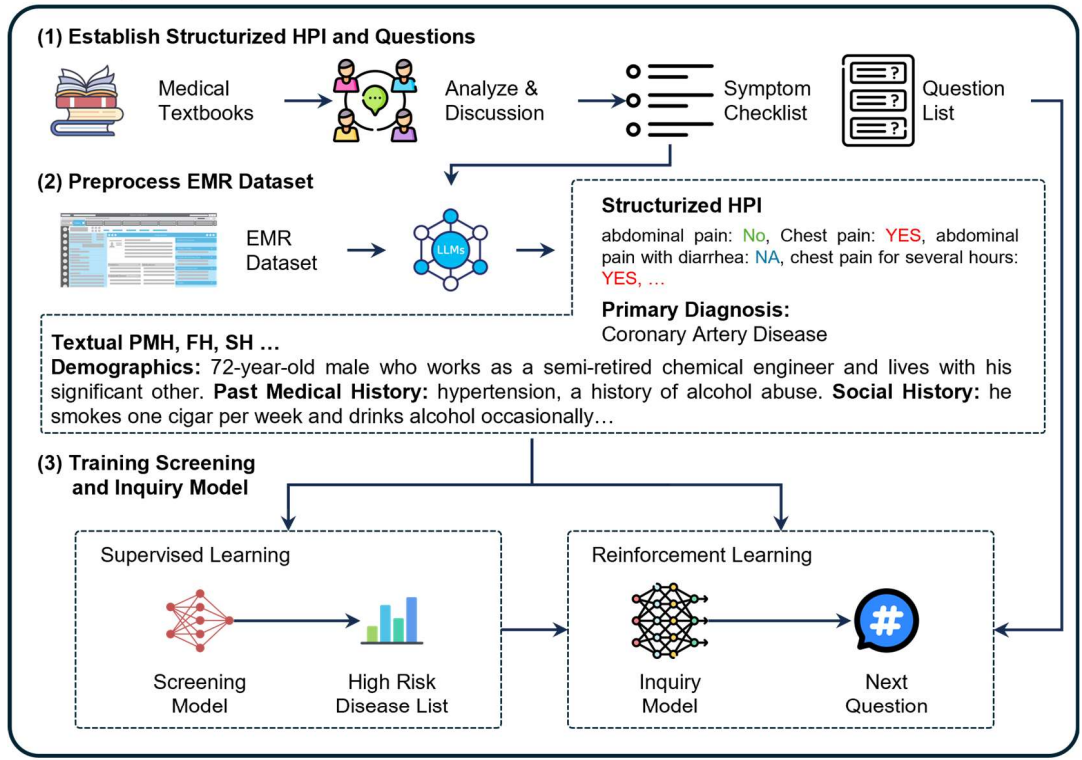
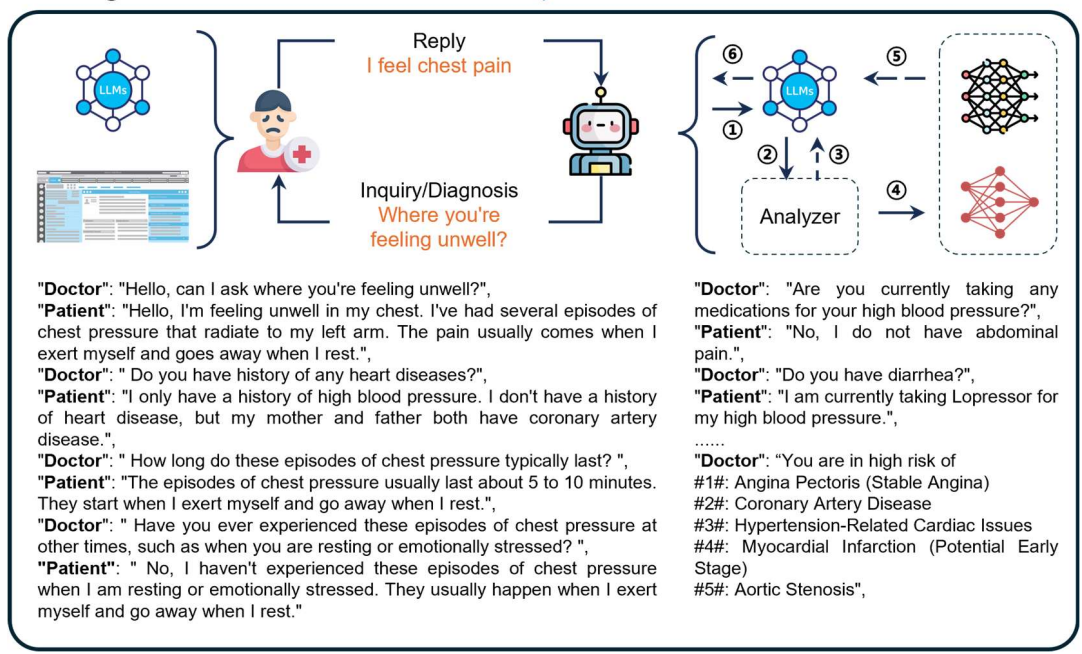

Figure 1 Workflow Pipeline of training and test PPME LLM.

We evaluated our plug-and-play model-enhanced (PPME) LLM agent through simulated clinical conversations. One LLM was configured to emulate patient responses based on EMR data, while another was set up to emulate doctors. The doctor LLM performed dialogue analysis using an analyzer component. The analyzer extracts structured HPI and other historical information through the conversation history at each turn, similar to the data preprocessing pipeline. The trained models then informed subsequent inquiry strategies and initial diagnoses based on the structured HPI and historical data. This hybrid architecture effectively leverages both the natural language processing capabilities of LLMs and the specialized functionalities of plugin models for medical inquiry and initial diagnosis. By implementing a plug-and-play approach instead of direct fine-tuning of LLMs, our methodology successfully circumvents the computational constraints typically associated with LLM optimization, enhancing its application potential.

Our experimental results indicate that the proposed approach effectively narrows the performance gap of LLMs in interactive diagnostic tasks. During the initial diagnosis phase, the PPME LLM demonstrated a significant performance boost of at least 30% compared to baselines. Utilizing the PPME LLM led to improved final diagnostic accuracy, closely aligning with results derived from using complete EMRs. These advancements suggest potential benefits for developing an autonomous diagnostic system and enhancing clinical decision support systems.

## 2 Result

### 2.1 Dataset Preprocessing

The data selection methodology for this study is depicted in Figure 2(a). We analyzed two datasets: the Sir Run Run Shaw Hospital (SRRSH) dataset from the School of Medicine, Zhejiang University, China, and the MIMIC dataset from Beth Israel Deaconess Medical Center, United States of America. The SRRSH dataset comprises admission EMRs of 3,196,836 visits (2013–2020), while the MIMIC dataset is the combination of MIMIC-III and MIMIC-IV datasets, which contain 384,519 admission EMRs from intensive care unit (ICU) admissions from 2001 to 2019 [25,26]. To ensure data quality, a three-stage protocol was implemented. First, EMRs with formatting errors or missing data were excluded. Second, diagnoses that occurred fewer

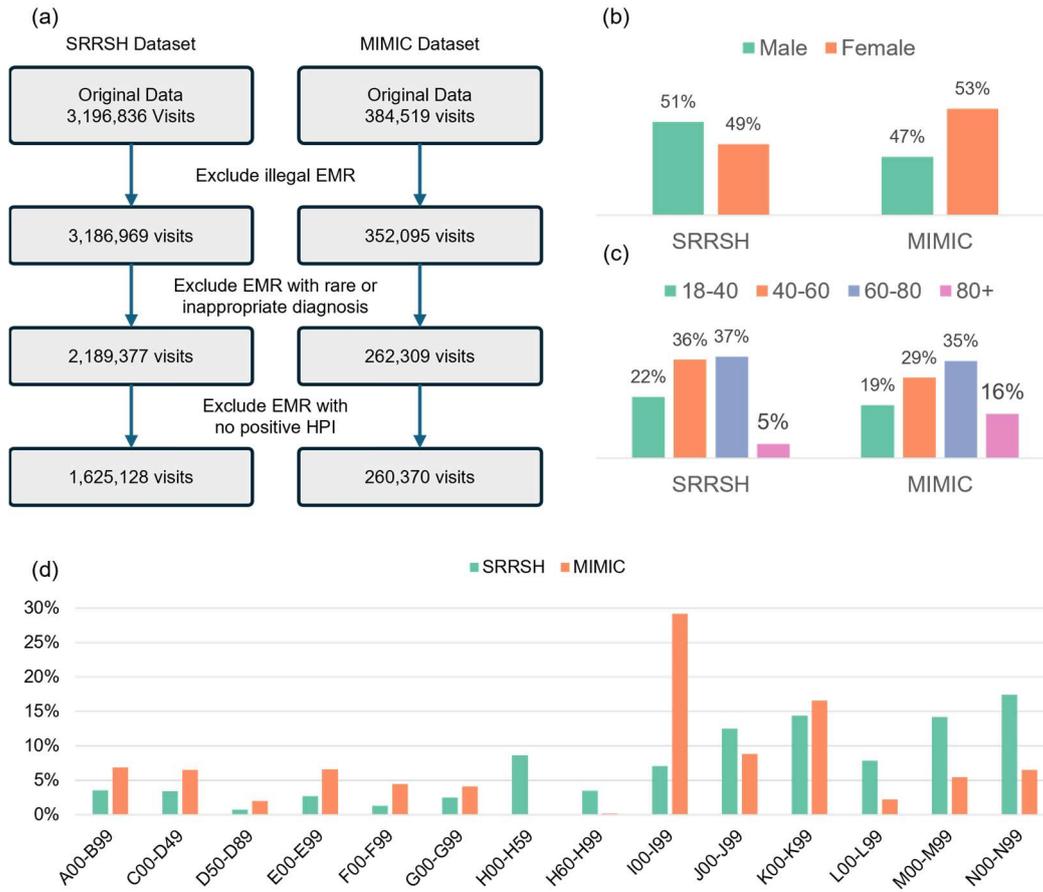

Figure 2 Data Description. (a) Data filtering pipeline. (b) Gender distribution. (c) Age Distribution. (d) Disease distribution (with ICD-10 Coding).

A00-B99 means certain infectious and parasitic diseases, C00-D49 means neoplasms, D50-D89 means diseases of the blood and blood-forming organs and certain disorders involving the immune mechanism, E00-E99 means endocrine, nutritional and metabolic diseases. F00-F99 means mental and behavioral disorders, G00-G99 means diseases of the nervous system, H00-H59 means diseases of the eye and adnexa, H60-H99 means diseases of the ear and mastoid process, I00-I99 means diseases of the circulatory system, J00-J99 means Diseases of the respiratory system, K00-K99 means diseases of the digestive system, L00-L99 means diseases of the skin and subcutaneous tissue, M00-M99 means diseases of the musculoskeletal system and connective tissue, N00-N99 means diseases of the genitourinary system.

than 20 times were removed to filter out rare conditions. We included EMRs with a primary diagnosis classified under ICD-10 (International Classification Disease) codes in the A00-N99 range. EMRs with ICD codes after O00 were excluded, as this range primarily encompasses conditions that are not traditionally classified as diseases (e.g., pregnancy, injuries, symptoms) or congenital disorders, which are beyond the scope of this study. Third, admissions lacking meaningful HPI information, such as routine follow-ups, were filtered. Finally, 1,625,128 admissions from SRRSH and 260,370 from MIMIC were retained.

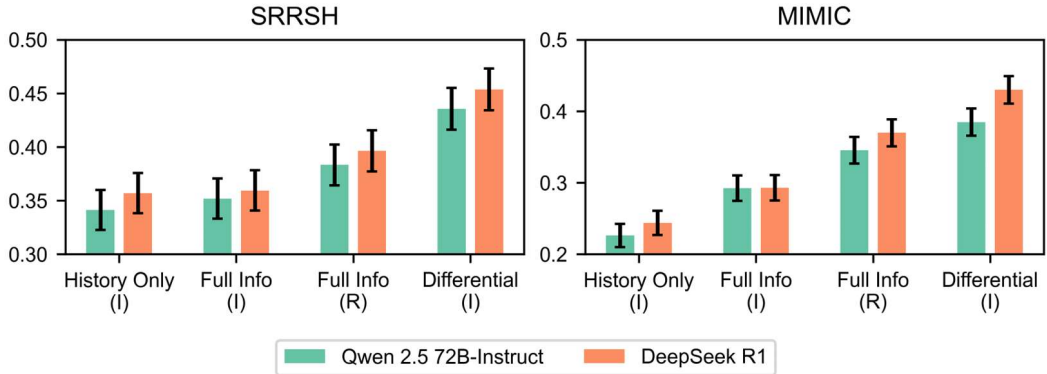

Figure 3. Phase-Specific Performance Evaluation. The figure indicates the ratio at which an LLM can generate correct diagnoses via interactive dialogue or review. Four settings were evaluated. "History only (I)" indicates that the LLM only inquiries about the medical history via interactive dialogue. "Full info (I)" indicates that the LLM can inquire about arbitrary information via interactive dialogue. "Full info (R)" indicates that the LLM can directly access complete EMRs without interactive dialogue. "Differential (I)" indicates that the LLM generates diagnoses via interactive dialogue according to a preset high-risk disease set. The error bar denotes the 95% confidence interval (CI).

Notable differences exist between the datasets. SRRSH encompasses outpatient and inpatient data across all medical departments, whereas MIMIC exclusively includes patients visiting the ICU. SRRSH shows a balanced male-female ratio and MIMIC has a higher proportion of female patients (Figure 2(b)). MIMIC patients are older on average than SRRSH patients, aligning with their ICU-specific focus (Figure 2(c)). Disease distribution in SRRSH is relatively balanced, but MIMIC exhibits significant skewness, with nearly 50% of admissions linked to circulatory and digestive disorders and minimal representation of eye or ear conditions (Figure 2(d)).

## 2.2 Phase-Specific Performance Evaluation

We conducted an evaluation of LLMs performance across various experimental conditions on 2,500 randomly selected samples in both datasets to investigate the specific factors affecting its planning capabilities (Figure 3). The analysis revealed that LLM exhibited its lowest interactive diagnostic performance when restricted to medical history queries. Inquiring about arbitrary information recorded in EMR results in minimal performance improvement in the SRRSH dataset. While the MIMIC dataset showed modest improvement with full EMR access in the interactive setting, the performance remained substantially below the review setting. A notable enhancement in performance occurred when the LLM was provided with a predetermined list of 10 high-risk diseases (including the correct discharge primary diagnosis) before conducting an interactive diagnosis. For example, in the MIMIC dataset, DeepSeek R1's diagnostic accuracy through interactive dialogue reached 0.43 when equipped with the high-risk disease list, demonstrating significant improvement compared to its performance with complete EMRs in review settings (0.38).

These findings suggest that the previously documented limitations in LLM's planning capability primarily stem from the bottleneck to conduct experience-based inquiry and initial diagnosis under incomplete information (i.e., the history only (I) setting). However, when presented with specific diagnostic objectives, the model demonstrates acceptable knowledge-based planning capabilities. Our subsequent work explored methods to improve this experience-based initial disease diagnosis capability.

## 2.3 Initial Diagnosis Internal Evaluation

We trained and evaluated the performance of plug-and-play models using two data sets, respectively. After that, we conducted dialogue-based medical consultations using 2,500 randomly selected cases from the testing set of each dataset, respectively. An example of an interactive initial diagnosis is recorded in Supplementary Figure 1. During a consultation, a doctor LLM agent was tasked to identify high-risk diseases in descending order after engaging in either 10 or 20 rounds of dialogue with the simulated patient agent. To assess consultation effectiveness, we employed the Recall@K metric. Our PPME LLM utilizes Qwen 2.5-72B-Instruct as the backbone LLM [27]. We conducted comparative analysis experiments using the current leading open-source general-purpose LLMs (Llama 3.3 70B-Instruct, Qwen 2.5 72B-Instruct), medical LLMs (UltraMedical 70B, OpenBioLLM 70B), and two reasoning-enhanced LLMs (HuatuoGPT o1 72B, DeepSeek R1) [3,27-31]. All baseline LLMs performed interactive initial diagnosis via their abilities.

Our experimental results demonstrate the effectiveness of using extensive real EMR data (Figure 4). The PPME LLM exhibited notably superior performance across both datasets (Figure 4(a)). Qwen 2.5 obtained the best performance among the baseline models in the SRRSH dataset experiment. However, our model demonstrated substantial performance improvements by equipping the PPME module. The Recall@1 through Recall@5 metrics increased from initial values of 0.34, 0.45, 0.50, 0.54, and 0.56 (Qwen 2.5) to enhanced values of 0.49, 0.65, 0.73, 0.78, and 0.81 (PPME LLM), respectively. The performance improved by approximately 40%. Analysis of the MIMIC dataset revealed lower overall model performance. This can be attributed to the dataset's exclusive focus on ICU cases, which typically involve complex medical conditions that are difficult to diagnose solely through patient history collection. The Llama 3.3 obtained the best performance among baselines, while our PPME LLM also obtained about 30% performance superiority (e.g., Recall@1 increased from 0.23 to 0.31). These results indicate that PPME LLM can generate more accurate initial diagnoses during interactive consultations with insufficient information.

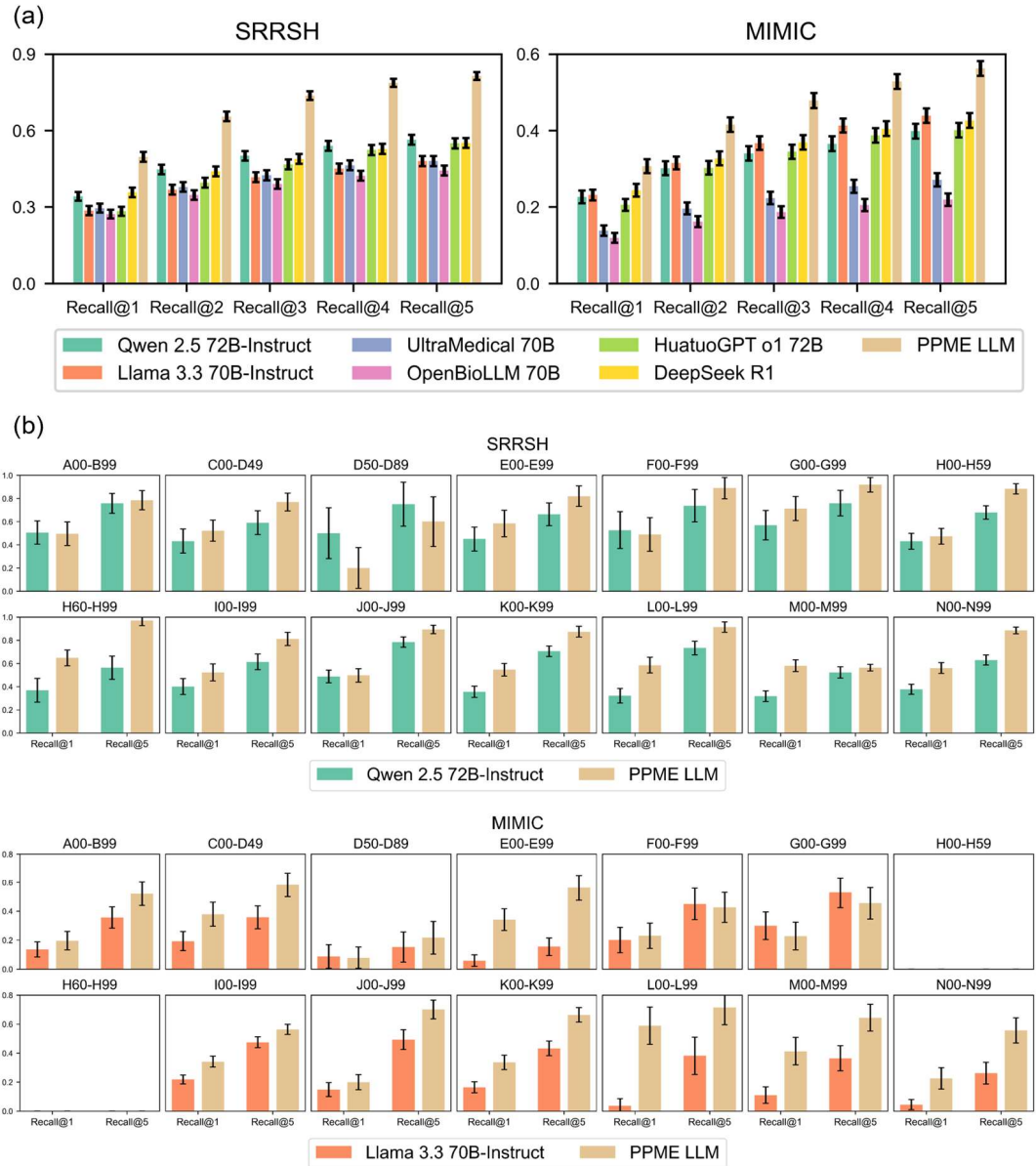

Figure 4 Internal Validation. (a) Diagnosis performance with 10 questions. (b) Detailed diagnostic performance per disease category. Due to the insufficient ophthalmic and ear disease data in the MIMIC dataset, the H00-H59 and H60-H99 categories are empty.

We found that the performance of two medical LLMs is the worst. In the SRRSH dataset, their performance was significantly lower than Qwen 2.5. For example, the Recall@1 through Recall@5 metrics in SRRSH decreased from 0.34, 0.45, 0.50, 0.54, 0.56 (Qwen 2.5) to 0.29, 0.38, 0.42, 0.46, 0.48 (OpenBioLLM), respectively. In the MIMIC dataset, their performance was significantly lower than all other models. The

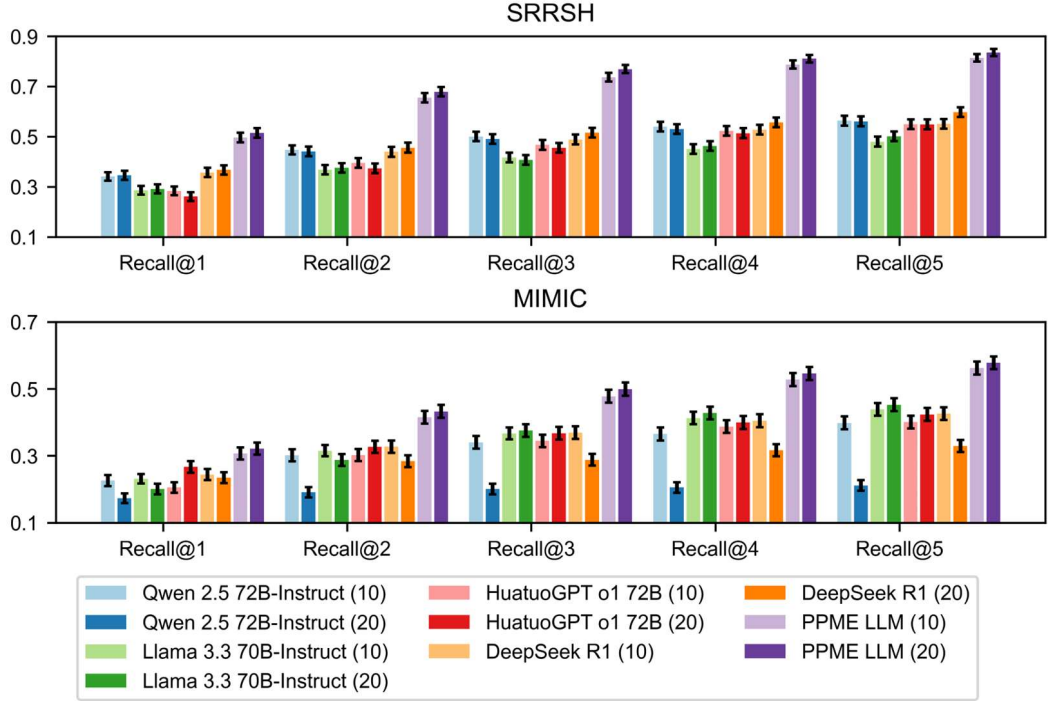

Figure 5 Performance differences with asking 10 and 20 questions.

Recall@1 through Recall@5 metrics decreased from 0.23, 0.32, 0.37, 0.41, 0.44 (Llama 3.3) to 0.14, 0.20, 0.22, 0.25 and 0.27 (OpenBioLLM), respectively. The performance of OpenBioLLM is comparable to UltraMedical. Consequently, we did not continue using these two models in subsequent experiments. This phenomenon suggests that although the fine-tuning process of medical LLMs improves their performance in medical question answering, it may potentially harm their planning abilities. This decrease may be attributed to insufficient exposure to medical interaction data during fine-tuning, resulting in the "catastrophic forgetting" or "alignment tax" effect [32] Meanwhile, we found that although reasoning LLMs take 5-10 times longer than other baselines to complete an initial diagnosis, there was no significant improvement in performance. For example, in the two datasets, DeepSeek R1's recall@1 to recall@5 performance was 0.36, 0.44, 0.49, 0.53, 0.55 (SRRSH), and 0.24, 0.33, 0.37, 0.40, 0.43 (MIMIC), respectively; while the best other baseline was 0.34, 0.45, 0.50, 0.54, 0.56 (SRRSH), and 0.23, 0.32, 0.37, 0.41, 0.44 (MIMIC). The performance of HuatuoGPT o1 is comparable to the DeepSeek R1.

In Figure 4(b), we evaluate the performance differences between PPME LLM and the best baseline model across different disease categories. The results show that PPME LLM achieves performance comparable to or significantly better than the optimal baselines in most disease subcategories. There are only two exceptions: in the SRRSH dataset, PPME LLM underperforms Qwen 2.5 in screening hematological diseases (ICD-10 D50-D89), and in the MIMIC dataset, it lags Llama 3.3 in screening neurological diseases (ICD-10 G00-G99). However, neither of these performance gaps reaches statistical significance. These findings suggest that PPME LLM demonstrates a relatively stable performance advantage across various disease subcategories.

In Figure 5, we further investigated the impact of asking more questions. We allowed each model to make an initial diagnosis after asking 20 questions and compared the performance after asking 10 questions. The experimental results show that none of the models achieved significant performance improvement by asking more questions. For example, the PPME LLM's recall@1 to recall@5 performance was 0.50, 0.66, 0.74, 0.79, 0.81 (SRRSH), and 0.31, 0.42, 0.48, 0.53, 0.56 (MIMIC) with 10 questions, respectively; while its performance was 0.51, 0.68, 0.77, 0.81, 0.84 (SRRSH), and 0.32, 0.43, 0.50, 0.55, 0.58 (MIMIC) when asking 20 questions. This phenomenon may indicate that asking 10 questions is an appropriate threshold of questions for the task of LLM-based interactive initial diagnosis. Therefore, in subsequent experiments, we conducted all experiments using 10 rounds of questions. On the other hand, we observed that many models have experienced performance decreasing when asking more questions. This may indicate that LLMs are more prone to hallucination during long conversations, a phenomenon that has also been observed in other related studies [14]. We also conducted external validation to examine the generalization performance of the PPME LLM, with the relevant experimental results presented in the supplementary material section 8.5.

## 2.4 Cross Analysis

We further investigated the origin of performance improvements via two experiments. In the first experiment, we reanalyzed dialogue records generated by both the LLM (Qwen 2.5) and the PPME LLM to evaluate how much of the HPI information documented in the EMRs could be rediscovered in the dialogues. The underlying hypothesis here is that if the plugin's inquiry strategy is superior, it should elicit more HPI information from the patient. In the second experiment, we used an LLM (Qwen 2.5) to infer a high-risk disease list based on dialogues generated by PPME LLM. This was done to analyze the performance difference between PPME LLM and the pure LLM when their inputs were entirely identical. This approach helps explore the performance gains of SL-trained models compared to zero-shot learning-based LLMs.

Figure 6 demonstrates that our plugin performs better in HPI inquiry strategies and initial diagnostic accuracy. In the SRRSH dataset, the PPME LLM obtained 0.82, 0.67, and 0.73, and Qwen 2.5 obtained 0.83, 0.60, and 0.70 in precision, recall, and F1, respectively. In the MIMIC dataset, the PPME LLM obtained 0.77, 0.60, and 0.67, and Qwen 2.5 obtained 0.80, 0.53, and 0.63 in precision, recall, and F1, respectively (Figure 6(a)). We found that the inquiry policy of PPME LLM outperformed the LLM in terms of inquiry recall, meaning it was noticeably better at identifying positive HPI documented in the EMR. However, the precision of HPI discovered by PPME LLM was comparable to or slightly worse than Qwen 2.5. The false positives in precision could be attributed to ambiguity or hallucinations in the LLM's responses while simulating a patient. Based on the F1 score analysis, the PPME LLM agent demonstrates superior overall performance compared to their LLM counterparts.

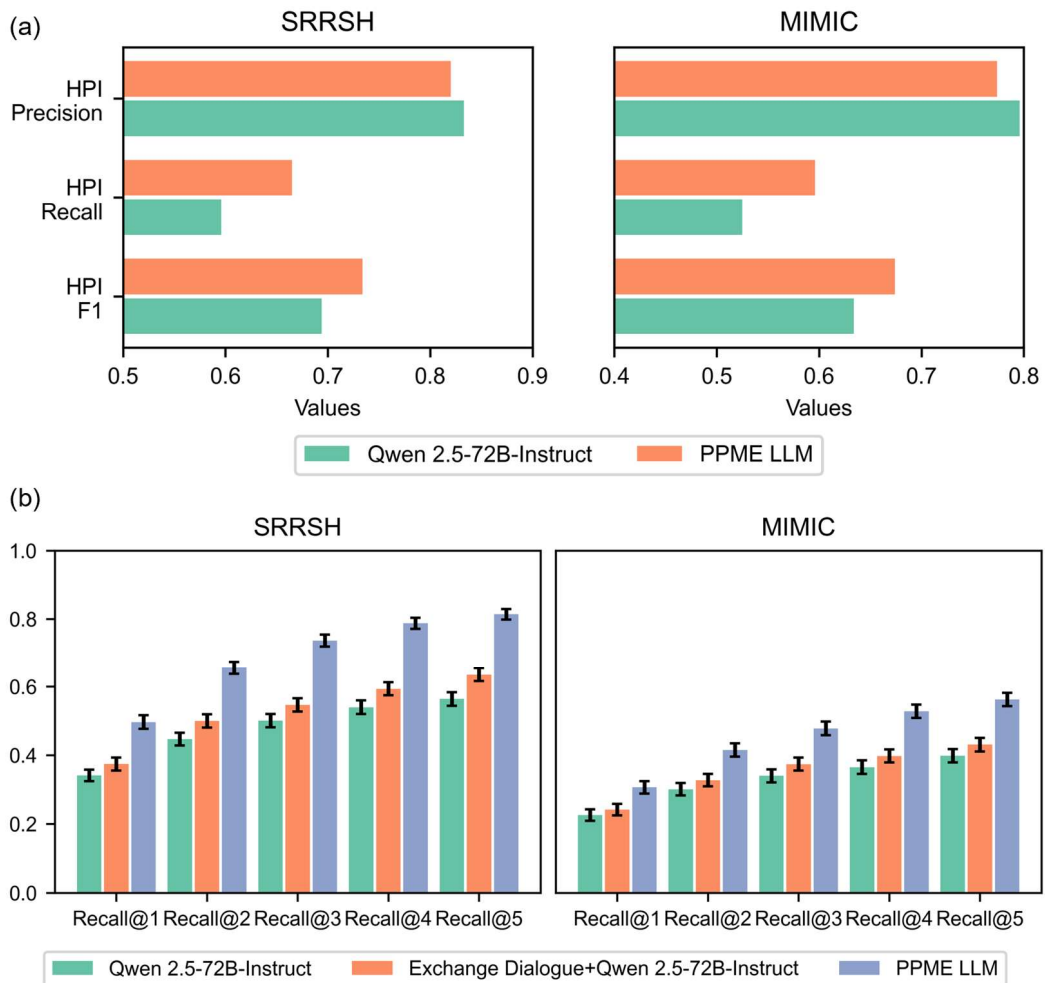

Figure 6 Cross Analysis. (a) HPI rediscovery analysis result. (2) dialogue exchange analysis result.

Figure 6(b) reveals that when Qwen 2.5 analyzes dialogues generated by PPME LLM agent performs re-diagnosis, increasing overall performance by approximately 6%-12%. The re-diagnosis Recall@1 to Recall@5 performance was 0.37, 0.50, 0.55, 0.60, 0.63 (SRRSH), and 0.24, 0.33, 0.37, 0.40, 0.43 (MIMIC), respectively, while its original performance was 0.34, 0.45, 0.50, 0.54, 0.56 (SRRSH), and 0.23, 0.30, 0.34, 0.37, 0.40 (MIMIC). This improvement suggests that the extra HPI obtained through plugin interactions positively influences the diagnostic capabilities of baseline LLMs. Based on these experimental results, approximately 25% of the performance improvement achieved by PPME LLM can be attributed to their ability to elicit more critical information.

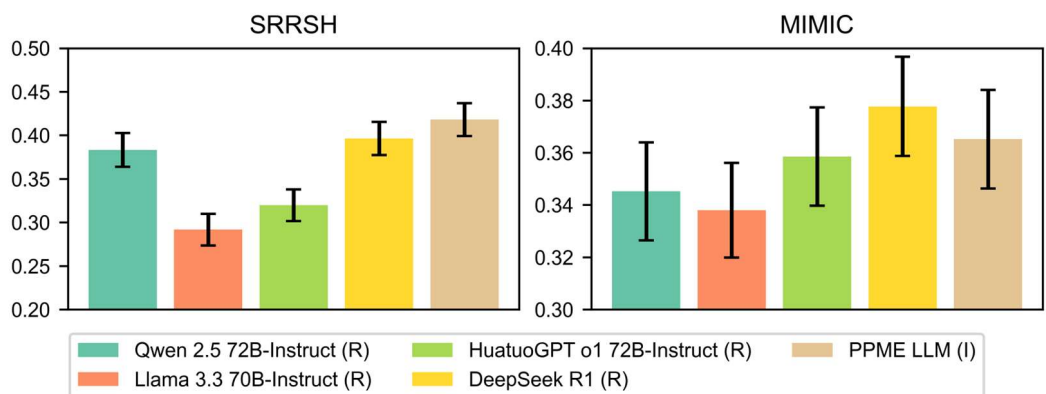

Figure 7 Final Diagnosis Analysis. "(I)" indicates that the LLM makes diagnosis via interactive dialogue. "(R)" indicates the LLM makes diagnosis by directly reviewing the entire admission EMR, without interactive dialogue.

## 2.5 Final Diagnostic Performance Evaluation

Finally, we evaluated the diagnostic capabilities of different LLMs (Figure 7). We used four leading open-source LLMs to review complete EMRs and provide their diagnostic conclusions. We also had PPME LLM perform the same task. The process began with ten rounds of question-answer interactions using PPME LLM to establish initial diagnoses, identifying the top 10 diseases with the highest risk. Subsequently, using DeepSeek R1, we conducted 10 rounds of interactive differential diagnosis dialogue based on these 10 diseases to generate the final diagnosis (Supplementary Figure 2).

Analysis of the experimental results revealed that DeepSeek R1 demonstrated superior performance among the baseline models, achieving diagnostic accuracy of 0.39 and 0.38 on the SRRSH and MIMIC datasets, respectively. These findings suggest that enhanced reasoning capabilities can improve diagnostic performance in the review setting. In the interactive diagnostic context, our proposed model exhibited a diagnostic accuracy 0.42 on the SRRSH dataset, surpassing DeepSeek R1's performance. Our model achieved a diagnostic accuracy of 0.36 on the MIMIC dataset. The model demonstrated marginally better performance than Qwen 2.5, Llama 3.3, and HuatuoGPT o1 while showing slightly lower performance than DeepSeek R1. However, these performance differences did not reach statistical significance. These results indicate that our approach has effectively addressed the performance deterioration problem of LLMs in interactive diagnostic scenarios.

## 3 Discussion

This study's primary clinical relevance lies in addressing the performance discrepancy of LLMs between single-round question-answering diagnostic tasks and interactive diagnostic tasks. Previous research has shown that LLMs can demonstrate diagnostic accuracy comparable to or exceeding that of human clinicians when provided with comprehensive information. However, the significant decline in performance during interactive consultations has limited LLMs' application perspective.

Our investigation reveals that the main challenge in LLM-based automated diagnosis arises from two key limitations: (1) the inefficient collection of medical history and (2) the inaccurate formulation of initial diagnoses when presented with incomplete information. The imprecise nature of initial diagnoses adversely affects the LLMs' ability to gather crucial differential diagnosis information later, decreasing their overall accuracy. Notably, when initial diagnoses include the actual diagnosis, LLMs demonstrate acceptable capability in diagnostic planning and can independently conduct differential diagnoses.

The findings suggest that improving the initial disease diagnosis capabilities of LLMs is crucial for enhancing their performance in the entire interactive diagnostic process. Our evaluation of specialized LLMs with enhanced reasoning capabilities, including DeepSeek R1 and HuatuoGPT o1, indicates no significant improvement in initial disease diagnosis performance. This observation leads to the hypothesis that effective disease diagnosis may rely more on experiential pattern recognition rather than solely on medical knowledge and logical reasoning. This near-intuitive experience is often regarded as a key advantage senior doctors have over junior doctors [33]. While recent advancements in LLMs have mainly focused on enhancing medical knowledge integration, reasoning capabilities, and reducing hallucinations, these improvements primarily benefit the differential diagnosis phase. However, these LLMs seem unable to accumulate the clinical experiences necessary for initial diagnosis since such experiences are not recorded in medical literature but are gained through years of hands-on clinical practice. This results in difficulty asking appropriate questions during experience-based, interactive initial consultations and challenges in making the most accurate inferences when information is limited. Current LLMs face hurdles in optimizing medical workflows due to regulatory requirements, patient privacy concerns, and computational constraints. These restrictions prevent them from accessing and learning from real clinical interaction data to emulate physician behavior. This fundamental limitation explains why existing LLMs often perform poorly in interactive diagnostic scenarios.

We developed a virtual environment that allows LLM agents to gain experience through simulated experiments, ultimately enhancing their performance in the initial disease diagnosis stage. We systematically outlined the information and related questions that can be asked during a medical consultation. By leveraging LLMs, we organized extensive real-world EMR datasets to create a dataset comprising numerous simulated patient cases within a discrete space. We employed RL and SL techniques to develop and optimize inquiry and diagnosis models through simulated medical consultations, similar to how physicians accumulate experience. Utilizing over 3.5 million medical EMR records to train the inquiry and diagnosis model, our agent achieved over a 30% performance improvement in both the SRRSH and MIMIC datasets, validating the significance of incorporating clinical experience. Specifically, in the SRRSH dataset, while the baseline LLM attained a recall@5 of 56%, the PPME LLM significantly raised this metric to 81%. This indicates that when using the LLM to recommend advanced medical examinations for further differential diagnosis based on suspected high-risk diseases, around 44% of cases might not reach a final diagnosis. In contrast, the PPME LLM may reduce this failure rate to 19%. This advancement could greatly enhance the allocation of medical resources while shortening the time needed for accurate diagnosis, ultimately leading to more efficient healthcare delivery. Following the screening of high-risk diseases through PPME LLM, the interactive diagnostic performance of LLMs became largely equivalent to their direct diagnostic performance based on complete medical records. This study's improvement can be attributed to the advantages of our SL and RL models over zero-shot approaches. Current LLMs are primarily trained in clinical literature and operate under zero-shot conditions when tackling clinical tasks.

The main technical value of this research lies in the accessibility of our method. We solely utilized patients' EMR data in our study. As healthcare institutions have amassed substantial EMR data over time, they can train targeted initial diagnosis models based on their proprietary data, specifically fitting their patient characteristics. We did not fine-tune the LLM, which provides three key benefits to this study. First, it enables healthcare institutions to develop and deploy tailored LLM agents with minimal hardware requirements, such as a single GPU card, significantly enhancing its applicability. Second, LLM fine-tuning often results in catastrophic forgetting, where enhancements in one domain lead to performance declines in others. For instance, in this study, we observed that current medical LLMs perform worse than general-purpose LLMs in interactive medical consultations, even as these medical LLMs achieve better results in medical LLM benchmarks. These performance declines are difficult to evaluate comprehensively; we aimed to avoid compromising other medical capabilities while improving interactive disease diagnosis performance [34]. Third, given the rapid advancements in LLM base models, we aimed to develop a solution independent of specific foundational LLMs. By proposing a plug-and-play method, we can seamlessly integrate our models with the latest LLMs as they are released.

Another advantage of this study is our use of a vast amount of real-world patient data from various healthcare institutions to evaluate the model's performance across a wide range of diseases. Our evaluation, conducted using thousands of real cases across hundreds of diseases from actual Chinese and American healthcare systems, demonstrated the agent's impressive and robust performance in interactive initial diagnosis tasks. To our knowledge, the number of real-world medical records and diseases included in this study's evaluation process is the highest among related studies.

However, as a proof-of-concept study, our research has several limitations. Since all studies were based on patients' admission EMR data, there may be missing information in this data segment, which could have affected the LLM's diagnostic performance, preventing it from reaching optimal levels. We did not incorporate physical examination information, as collecting such data via dialogue is impractical. Although experimental results indicate that our PPME LLM has demonstrated greater efficiency in information inquiry, approximately 30% of the HPI elements identified by physicians during real medical consultations were undiscovered. This suggests significant room for improvement in our model. While our study emphasized efficient disease identification in history collection, it did not explore the comprehensiveness of the information-gathering process. Future research will need to include clinical protocols to ensure more thorough history-taking.

## 4 Method

### 4.1 LLM Selection and Deploying

Due to privacy protection regulations, experiments in this study were conducted based on open-source LLMs that can be deployed locally. Unless otherwise specified, all experiments were conducted using the Qwen 2.5 72B-Instruct [27]. Qwen 2.5 72B-Instruct is one of the leading open-source LLMs and has achieved performance comparable to GPT-4o in major benchmarks [35,36]. We also employ the latest general and medical LLMs for comparison. The vLLM framework was utilized to accelerate LLM inference speed [37].

### 4.2 Dataset and Preprocessing:

We developed two datasets for training and validation. The first training dataset is the SRRSH dataset, which contains admission of hospitalized and outpatient patients from the Sir Run Run Shaw Hospital, School of Medicine, Zhejiang University, Hangzhou, China. This dataset contains a complete Chinese EMR of 3,186,836 admissions covering all hospital departments. The second dataset is the MIMIC dataset, which contains admissions from the ICU of Beth Israel Deaconess Medical Center, Boston, MA. This dataset combines MIMIC-III and MIMIC-IV datasets, which contain 384,519 English patient EMRs [25,26]. Both datasets used the ICD system to record diagnosis. Our study was performed under research protocols approved by the institutional review boards (IRB) at Zhejiang University. For both datasets, IRB approved a waiver of consent as they were de-identified before access and were performed in accordance with a research data use agreement. There was no direct interaction with any patients at any point during this study.

We applied a unified data preprocessing method to the training dataset. We first mapped all ICD-9 diagnostic codes to ICD-10 and retained only the first three digits of the ICD code [38]. Diseases with the same first three digits of ICD codes were considered identical, with no further distinction. The first diagnosis of patients was used as the target diagnosis label. We then reserved only those ICD codes within the range A00-N99 and those with more than twenty occurrences, excluding rare diseases from the scope of prediction. Diseases starting with the code O00 were excluded from the diagnostic process as they primarily refer to congenital diseases, pregnancy, symptoms, and trauma, which were beyond the scope of this study. We employed a rule-based method to preprocess EMR documents, retaining only the patient's chief complaint, history of present illness, past medical history, social history, and family history. Even though the physical examination information is important for diagnosing, it was excluded. We finally filtered admission records and did not report any positive HPI elements, primarily excluding patients undergoing routine follow-ups.

Finally, the SRRSH dataset contains 1,625,128 records with an average EMR length of 248 words. The MIMIC dataset contains 260,370 admissions with an average EMR length of 1048 words. The two datasets contain 512 different diseases. After randomization, we split the two datasets into training, validation, and test sets in a ratio of 0.6, 0.1, and 0.3, respectively.

## 4.3 Model Training

### 4.3.1 Training Phase Simulated Patient

The simulated patient dataset for training $\mathcal{D} = \{h_i, p_i, y_i\}_{i=1}^N$ contains $N$ different simulated patient samples. Each sample consists of three parts. $h_i$ is the textual demographic information, past medical history, family history, and personal history of $i$-th patient, which is extracted from EMR via an LLM. $p_i \in \{0,1,2\}^M$ encodes structured HPI, where $M$ denotes the number of HPI elements, and zero, one, two denotes HPI elements are not mentioned, confirmed, or denied, respectively. $y_i$ denotes the primary discharge diagnosis.

This design reflects the differing characteristics of information collection in medical history taking. Collecting past medical history, family history, and personal history often faces challenges such as missing medical records, patient forgetfulness, or intentional concealment. Even experienced physicians find it difficult to address these issues, which fall outside the scope of this study. This study assumes that such information is available and can be obtained through straightforward questioning during consultations without relying on complex techniques. Therefore, these types of information are directly recorded as natural text. The simulated patient will return the corresponding answer when a relevant question is asked. The primary challenge in collecting information on current medical history lies in the fact that patients are often unable to provide a complete description of their HPI [39]. As patients may exhibit a subset of hundreds of possible HPI elements, it is also impractical to exhaustively query every potential HPI element. Doctors need to identify positive HPI elements via their own experience with only several questions, thereby facilitating the formulation of a comparatively accurate diagnosis. Therefore, it is necessary to employ a structured approach to convert all possible HPI into binary vectors, simplifying the diagnostic interaction from natural language to interactions within discrete state-action spaces, which can subsequently support the training of inquiry strategies.

We adopted a two-level structure to describe HPI, which is widely used in existing symptom checkers [40,41]. The first-level HPI is used to describe categories (e.g., chest pain), while second-level HPI provides more detailed descriptions (e.g., chest pain was radiating, chest pain lasting from several minutes to several hours, or chest pain accompanied by fever). The construction of the HPI list in this study was carried out in two stages. In the first stage, three doctoral students in clinical medicine or students with master's degrees in clinical medicine from Chinese medical schools were recruited to review standard Chinese medical textbooks manually. They independently develop comprehensive HPI lists of all diseases mentioned. After that, they conducted internal discussions and summarized an initial HPI elements list, resolving ambiguous HPI elements through a voting process. In the second stage, the preliminary HPI elements list was reviewed and refined by two licensed clinical physicians and one PhD researcher specializing in medical artificial intelligence. During this stage, the research team also consulted existing HPI elements checkers to finalize the construction of the HPI elements list. The final HPI elements list comprised 85 first-level HPI elements and 1,177 associated second-level HPI elements. The details about the HPI list and question list can be found in supplementary materials 8.2.

Then, we attempted to enumerate all reasonable questions for inquiring about these HPI, totaling 1,398 questions. Among them, 1,264 are closed-ended questions that can be answered with "yes" or "no," each corresponding to one HPI element (e.g., "Are you feeling dizzy?"). Additionally, we summarized 134 open-ended questions (e.g., "Can you specify where exactly your stomach hurts?"). These questions correspond to several second-level HPI elements (e.g., upper abdominal pain, lower abdominal pain) to guide patients in actively describing their condition more practically during consultations.

We used an LLM to analyze EMRs and structurize HPI elements for each admission. The LLM is instructed to identify whether each first-level HPI was confirmed, denied, or not mentioned within an EMR. If a first-level HPI was identified as "denied" or "not mentioned," all its associated second-level HPIs were also marked as "denied" or "note mentioned." After processing all first-level HPI, the LLM was instructed to parse the second-level HPI, whose associated first HPI elements were confirmed. According to the procedure, we converted textual HPI into binary vectors for each EMR. We randomly selected 400 structured HPIs for human verification. The LLM demonstrated high accuracy in extracting HPIs, achieving both Recall and Precision rates of 97%. An example of HPI structurized result is recorded in the supplementary material 8.1. We also indirectly evaluated the comprehensiveness of the structured HPI, whose result is in Supplementary Figure 6.

### 4.3.2 Initial Disease Diagnostic Model Training

We adopted $\mathcal{D}$ to train a initial disease diagnosis model. The model is a simple feedforward neural network (FNN) with two hidden layers whose input is the concatenation of $e_i$ and $p_i$, where $e_i$ is the textual embedding of $h_i$ generated by an embedding model [42], and output is the predicted probability of each disease. The model can be optimized via a standard supervised learning manner. The model can generate the risk probability of each disease.

$$\hat{y}_i = \text{FNN}([e_i, p_i]) \tag{1}$$

$$\mathcal{L} = \sum_{i=1}^{N} \text{CrossEntropy}(y_i, \hat{y}_i) \tag{2}$$

### 4.3.3 Inquiry Model Training

The inquiry model is an FNN with two hidden layers that generate question topics and are trained using RL methods.

**State Space.** The space is represented as a vector $s_i^t \in N_+^{3M}$, where each HPI element is encoded as [1,0,0], [1,0,0], or [1,0,0], representing the element being currently unknown, confirmed, or denied, respectively. At the beginning of each episode ($s_i^0$), the positive first-level HPI, negative first-level HPI, positive second-level HPI, and negative second-level HPI are disclosed with probabilities $p_{1p}, p_{1n}, p_{2p}, p_{2n}$, respectively. Of note, the positive second-level HPI will be disclosed only when the corresponding first-level HPI is disclosed. This simulates the characteristic that patients proactively disclose their HPI at the beginning of a medical consultation.

**Action Space.** The action space $\mathcal{A}$ contains $K$ actions, where each question asks whether the related HPI elements are present in the next turn. We presume that there is also a two-layer structure within the action space. The first layer action asks for a first-level HPI, while a secondary layer action asks for more specific HPI (such as chest pain accompanied by fever). Each second-layer action is affiliated with a first-layer HPI. We stipulate that the model can only select second-layer actions after the affiliated first-layer HPI elements are confirmed. Meanwhile, we do not allow the model to ask the same question twice.

**Agent.** The agent in this study is an FNN with two hidden layers, which outputs the action to be predicted in the next round based on the state $s_i^t$ and $e_i$.

$$a_i^t = FNN([e_i, s_i^t]), a_i^t \in \mathcal{A} \tag{3}$$

**Environment.** The environment in this study is the simulated patient dataset $\mathcal{D}$. At the beginning of each episode, the environment randomly selects a patient as the current simulated patient. In each round, the environment receives a question from the agent, which inquires about the presence or absence of one or several HPI elements. The environment then provides the agent with information on the corresponding HPI elements.

Reward. We set the reward $R_t$ as:

$$R_t = -\lambda + \alpha(F_{1p} + \beta F_{1n}) + F_{2p} + \beta F_{2n} + \sum_j |\hat{y}_j^{t-1} - \hat{y}_j^t|, \qquad (4)$$

Where $\lambda$ is the time penalty coefficient, the $F_{1p}, F_{1n}, F_{2p}, F_{2n}$ denotes the number of newly identified first-level positive HPI elements, first-level negative HPI elements, second-level positive HPI elements, and second-level negative HPI elements in the current round. $\alpha, \beta$ are coefficients. We additionally evaluated the absolute difference in predictions of the initial disease diagnosis model between two consecutive dialogue rounds as part of the reward function. This design is inspired by an empirical observation during the experiments that the performance of the initial disease diagnosis model improves when more HPI information is obtained. Therefore, we assume that the greater the difference introduced to the output vector by the newly collected HPI information, the more significant the collected HPI is in improving the model's performance. We incorporated this signal into the reward function to guide the model in asking questions most beneficial for enhancing diagnostic performance.

**Optimization Method.** The parameters within this study's agent are optimized using the Proximal Policy Optimization (PPO) algorithm [43]. An additional independent FNN is utilized as the value network to assist in the optimization process.

## 4.4 Evaluation

### 4.4.1 Initial Diagnosis

We evaluated the diagnostic performance of LLM agents and PPME LLM agents through simulated multi-turn dialogues between simulated doctors and simulated patients. In compliance with existing regulatory requirements, we did not evaluate proprietary LLMs such as ChatGPT and Claude. We implemented a prompt of simulators aligning with recent research practices [12].

The plug-in enhanced doctor simulator consists of an LLM, the analyzer, the initial disease diagnostic model, and the inquiry model. The analyzer is a prompt-based component. During each conversation round, the analyzer utilizes an LLM to identify whether the patient has already reported their medical history, family history, and other relevant information in the existing dialogue. If such information has not been provided, the analyzer will instruct LLM to prioritize asking for this information (i.e., step 3 in Figure 1). If all such information has been collected, the analyzer will generate embeddings of this information using a text embedding model, denoted as $\hat{e}_i^t$. The analyzer also structures the HPI $\hat{s}_i^t$ from the historical dialogues in each round following the same process as data preprocessing. The inquiry model determines the $a_i^t$ via $\hat{e}_i^t$ and $\hat{s}_i^t$, which is the topic of the next round question (i.e., the step 4, 5 in Figure 1). Then, the LLM generates text-form questions via $a_i^t$. In each round, the patient simulator must answer the questions posed by doctor simulators based on the EMR. After preset $L$ round of conversations, the disease screening model generates $\hat{y}_i^L$ based on the latest $\hat{e}_i^L$ and $\hat{s}_i^L$.

The LLM-based agent utilizes a similar decision procedure, relying on itself to determine questions and the final diagnosis. After the conversation concludes, a DeepSeek V3-based LLM moderator compares the diagnosis provided by the doctor's simulator with the actual diagnosis recorded in the EMR to determine whether the diagnosis is correct [44]. We evaluated the moderator's performance in Supplementary Figure 5.

We adopted two validation approaches: internal validation and external validation. In the internal validation, we randomly selected 2,500 EMRs from the test datasets of two datasets to construct 2,500 simulated patients for testing. In the external validation, we conducted a cross-experiment test. Specifically, we extracted all ICU patient data from the SRRSH dataset to retrain the model. Subsequently, we randomly selected 2,500 ICU records from the MIMIC test dataset and tested them using the model trained on the SRRSH dataset and vice versa. When conducting cross-external experiments, the parsed textual medical history was translated before input into models.

We evaluate performance using the Recall@K rate of the real primary diagnosis in the ranking. For example, a Recall@3 rate of 0.50 means that in 50% of test cases, the true primary discharge diagnosis appears within the top three predictions generated by the disease screening planner.

4.4.2 Final Diagnosis

In Figure 3, we first input the patient's EMR into the PPME LLM to infer the probability of diseases. If the patient's actual diagnosis appears among the top 5 highest-risk diseases, these top five are directly used as the high-risk disease list. If the patient's actual disease is not among the top five, we compile a high-risk disease list consisting of the top four diseases plus the actual diagnosis. Subsequently, the LLM conducts interactive diagnosis based on the high-risk disease list and generates a diagnostic result.

In Figure 7, we employ PPME LLM to conduct a 10-round interactive diagnosis process to generate a high-risk disease list. If the actual disease does not appear within the top 10 positions on the list, the diagnosis is directly considered unsuccessful. If the actual disease is among the top 10, the LLM proceeds to ask ten questions to perform differential diagnosis among these 10 diseases. We then evaluate whether the LLM can identify the true diagnosis among all cases.

# 5 Data Availability

The raw data of the MIMIC dataset can be found at https://physionet.org/, but access is required due to its terms of use. Due to privacy regulations, the EMRs cannot be made freely available in a public repository. De-identified data from the SRRSH dataset can be requested through the management team by contacting the corresponding author, following a defined protocol for data request approval. Generally, all such requests for access to EMR data will be responded to within 1 month. Data can be shared only for non-commercial use.

# 6 Source Code Availability

The source code of the project can be found in:

https://github.com/DanielSun94/ppme

# 8 Supplementary Materials

## 8.1 Data Structurize Example

Please see the mimic_iv-hospitalization-10089119_22582998_detail.csv and mimic_iv-hospitalization-10089119_22582998_symptom.csv in the Github Project.

## 8.2 HPI and Question List

Please see the "HPI.csv" and "Question.csv" in the Github Project.

## 8.3 Interactive Initial Diagnosis Example

**Round 1:**
Doctor: What do you feel uncomfortable with right now?
Patient: I'm feeling short of breath, doctor. It's been getting worse over the last few days, and I can't even speak in full sentences without having trouble catching my breath.
**Round 2:**
Doctor: Thank you for sharing that. May I ask your age, gender, past medical history, and any family history of illnesses?
Patient: I'm a 67-year-old female. I have a history of atrial fibrillation, mitral valve prolapse. My family history includes a brother who passed away from a heart attack, and my mother and brother had palpitations. My sister also has mitral valve prolapse.
**Round 3:**
Doctor: I see. Can you tell me if you have any swelling in your feet or legs?
Patient: Yes, my legs have been swollen for about two weeks now, and I've noticed some fluid leaking from them.
**Round 4:**
Doctor: Have you been experiencing any chest pain or tightness along with the shortness of breath?
Patient: No, there's no chest pain or tightness, but I do feel some palpitations from time to time. I haven't had any chest pain either.
**Round 5:**
Doctor: That's helpful. Do you feel any irregular heartbeats when you experience shortness of breath?
Patient: No, I haven't noticed any unusual heartbeats. My breathing feels difficult, but my heart doesn't seem to be racing or skipping beats.
**Round 6:**
Doctor: I see. Are you coughing at all?
Patient: I don't have a cough.
**Round 7:**
Doctor: Have you had any limb pain?
Patient: Yes, I've been feeling some pain in my limbs, and my legs also feel cold.
**Round 8:**
Doctor: Are you experiencing any dizziness?
Patient: No, I haven't felt dizzy or lightheaded, just short of breath and a bit of discomfort from the swelling in my legs.
**Round 9:**
Doctor: When does your shortness of breath usually occurs?
Patient: It mostly happens at night, especially when I'm trying to sleep. The breathlessness keeps me awake.
**Round 10:**
Doctor: Have you experienced any nausea or vomiting?
Patient: No, I haven't had nausea or vomiting, but I feel more fatigued than usual. I'm exhausted from struggling to breathe and dealing with the swelling in my legs.

**Initial Diagnosis:** Heart Failure (I50), Atrial Fibrillation and Flutter (I48), Rheumatic Heart Disease (I09), Nonrheumatic Mitral Valve Disorders (I34), Hypertensive Heart Disease (I11), Pulmonary Embolism (I26), Pleural Effusion (J91), Acute Myocardial Infarction (I21), Hypertensive Heart and Chronic Kidney Disease (I13), Acute and Subacute Endocarditis (I33).

Supplementary Figure 1 Interactive Initial Diagnosis Sample

# 8.4 Interactive Differential Diagnosis Example

**ROUND: 1**
DOCTOR ASK: "Can you provide the results of your most recent blood test, including levels of BNP or NT-proBNP, as these can help assess heart function?",
PATIENT RESPONSE: My most recent blood test showed a ProBNP level of 6289.

**ROUND: 2**
DOCTOR ASK: "Have you had an echocardiogram recently? If so, could you share the findings, particularly regarding ejection fraction and any valve abnormalities?",
PATIENT RESPONSE: Yes, I had an echocardiogram recently. The findings showed an ejection fraction of 30%, which is lower than normal. There were also valve abnormalities, including severe aortic stenosis, moderate mitral regurgitation, and moderate to severe tricuspid regurgitation.

**ROUND: 3**
DOCTOR ASK: "Have you experienced any recent episodes of chest pain, shortness of breath, or swelling in your legs, and if so, can you describe the severity and duration of these symptoms?"
PATIENT RESPONSE: Yes, I have experienced shortness of breath, which has gotten worse over the past few days. It became so bad that I couldn't speak in full sentences. I also noticed swelling in my legs about two weeks ago, which was followed by fluid leaking from them. I haven't had any chest pain.

**ROUND: 4**
DOCTOR ASK: "Have you had a recent chest X-ray or CT scan, and if so, could you share the findings, particularly regarding any signs of fluid in the lungs or pleural space?",
PATIENT RESPONSE: Yes, I had a chest X-ray recently. The findings showed small to moderate-sized bilateral pleural effusions, meaning there is fluid in the pleural space around the lungs. There were also bibasilar patchy opacities, which could be due to atelectasis, infection, or aspiration. A follow-up X-ray after thoracentesis showed a decrease in the right pleural effusion, but the left pleural effusion remained stable.

**ROUND: 5**
DOCTOR ASK: "Have you had a recent electrocardiogram (ECG)? If so, could you share the findings, particularly regarding any abnormalities in heart rhythm or electrical activity?",
PATIENT RESPONSE: Yes, I had a recent ECG. The findings showed V-pacing with LBBB, and the ventricular rate was 80. There were no other abnormalities noted in the heart rhythm or electrical activity.

**ROUND: 6**
DOCTOR ASK: "Have you had a recent D-dimer test or CT pulmonary angiography to assess for any signs of pulmonary embolism?",
PATIENT RESPONSE: No, I have not had a recent D-dimer test or CT pulmonary angiography.

**ROUND: 7**
DOCTOR ASK: "Have you had a recent assessment of your kidney function, such as a blood test for creatinine or a urine test for proteinuria, to evaluate for any signs of chronic kidney disease?",
PATIENT RESPONSE: Yes, I had a recent blood test for creatinine. The results showed a creatinine level of 0.8, which is within the normal range. I have not had a urine test for proteinuria.

**ROUND: 8**
DOCTOR ASK: "Have you had a recent assessment of your blood pressure, and if so, could you share the readings, particularly any patterns of hypertension or elevated levels?",
PATIENT RESPONSE: Yes, I had a recent assessment of my blood pressure. The readings were 106/57 mmHg during admission and ranged from 115/36 to 127/72 mmHg at discharge. There were no specific patterns of hypertension noted; the readings were mostly on the lower side.

**ROUND: 9**
DOCTOR ASK: "Have you had a recent assessment of your liver function, such as a blood test for liver enzymes (AST, ALT) or imaging studies, to evaluate for any signs of hepatic congestion or dysfunction?",
PATIENT RESPONSE: No, I have not had a recent blood test for liver enzymes or imaging studies to evaluate my liver function.

**Round 10**
Based on the information provided, the final diagnosis is **#1: Heart failure**.
### Rationale:
1. **Elevated ProBNP (6289 pg/mL)**: This is significantly elevated and strongly indicative of heart failure.
2. **Ejection Fraction (30%)**: This is severely reduced, consistent with systolic heart failure.
3. **Valve Abnormalities**: Severe aortic stenosis, moderate mitral regurgitation, and moderate to severe tricuspid regurgitation contribute to heart failure.
4. **Symptoms**: Shortness of breath (worsening), leg swelling, and fluid leakage are classic symptoms of heart failure.
5. **Chest X-ray Findings**: Bilateral pleural effusions and bibasilar opacities are consistent with heart failure.
6. **Blood Pressure**: Low-normal readings are consistent with advanced heart failure, where the heart cannot maintain adequate perfusion pressure.
7. **ECG Findings**: V-pacing with LBBB aligns with heart failure management.
While other conditions like pleural effusion (#7) and nonrheumatic mitral valve disorders (#4) are present, they are secondary to the primary diagnosis of heart failure.

Supplementary Figure 2 Interactive Differential Diagnosis Example. DeepSeek R1 conducted differential diagnosis according to the top 10 high-risk diseases generated by the PPME LLM in Supplementary Figure 1.

## 8.5 External Validation

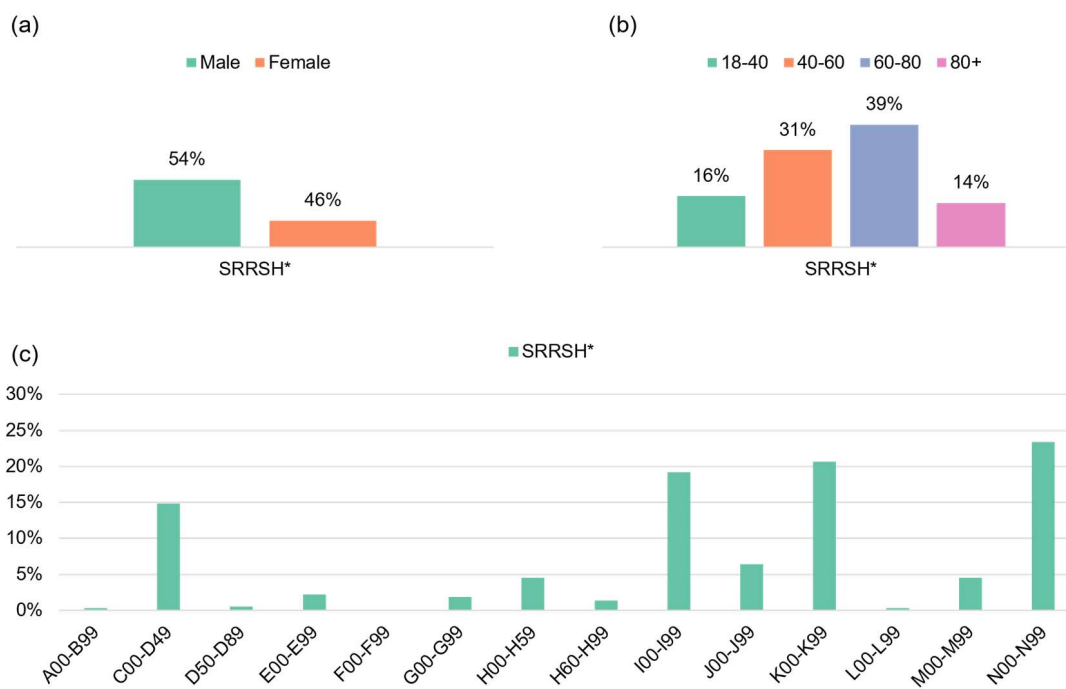

Supplementary Figure 3 SRRSH* data description. (a) Gender distribution. (b) Age Distribution. (c) Disease distribution (with ICD-10 Coding).

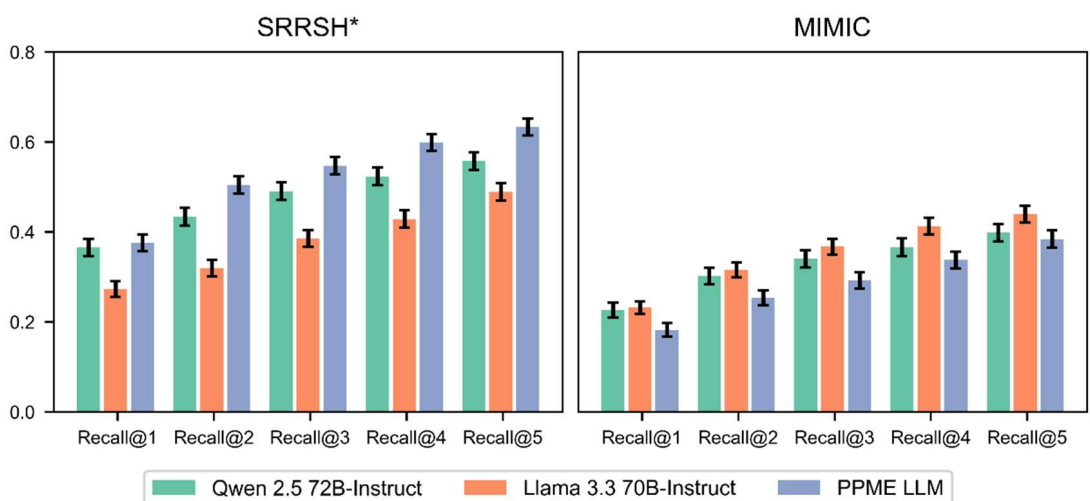

Supplementary Figure 4 External Validation

  We constructed the SRRSH* dataset for external validation. SRRSH* is a subset of the SRRSH dataset. Hospitalized cases meeting any of the following three criteria were included in SRRSH*: (1) explicit documentation of ICU admission in medical records; (2) receipt of surgical procedures; (3) presence of resuscitation records. Applying these three criteria, we identified 264,178 relevant records. Age distribution, gender distribution, and disease distribution of the cohort are shown in Supplementary Figure 3. Then, we conducted cross-external validation. Specifically, we used a model trained on the MIMIC dataset to test performance on the SRRSH* dataset (and vice versa).

The experimental results are shown in Supplementary Figure 4. Although the SRRSH* and MIMIC datasets originate from different countries, are written using different languages, and have significant differences in disease distribution, there is still some external generalization potential. For example, when using a diagnostic agent trained on the MIMIC dataset to make diagnoses on SRRSH*, its Recall@1 to Recall@5 performance is 0.38, 0.50, 0.55, 0.60, 0.63, which is better than the performance of Qwen 2.5 (0.37, 0.43, 0.49, 0.52, 0.56). When the PPME LLM trained on the SRRSH* dataset is used to train on the MIMIC dataset, its Recall@1 to Recall@5 performance is 0.18, 0.25, 0.29, 0.34, 0.38, which is slightly worse than the performance of Llama 3.3 (0.23, 0.32, 0.37, 0.41, 0.44). This result suggests that our PPME LLM model generalizes well and has the potential to be used in different healthcare facilities.

## 8.6 Moderator Performance

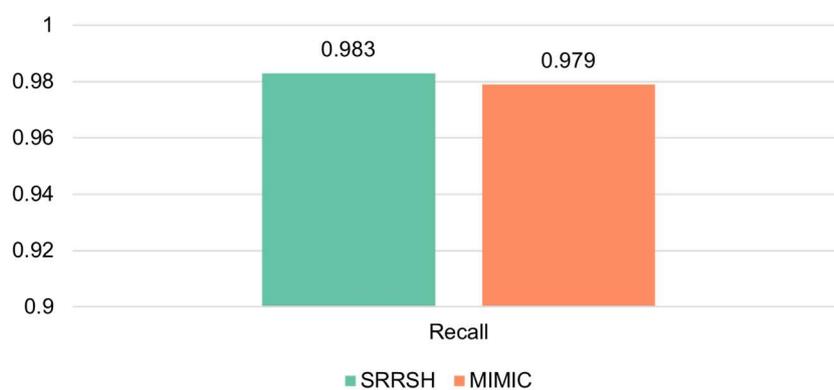

Supplementary Figure 5 Accuracy of LLM-based Diagnosis Moderator

In this study, all baseline LLMs generate diagnoses in natural text format through multi-turn dialogues. These diagnoses often show partial discrepancies with ICD codes. We utilized DeepSeek V3 as the moderator to automatically assess whether divergent natural language descriptions refer to the same disease. To validate DeepSeek V3's performance, we randomly sampled 10,000 EMRs from the SRRSH and MIMIC datasets, respectively. We extracted physicians' natural-text primary diagnoses (as documented in EMRs) alongside their ICD code counterparts and applied DeepSeek V3 for consistency evaluation. If DeepSeek V3 successfully identifies both descriptions as representing the same disease in most cases, this capability demonstrates its validity for implementation as a moderator. The results confirm robust disease-matching capabilities in DeepSeek V3 (Supplementary Figure 5). In the SRRSH dataset,

DeepSeek V3 achieved 98.3% agreement between ICD-coded diseases and physician-written diagnostic descriptions, with 97.6% accuracy in the MIMIC dataset. This validates its reliability as a moderator.

## 8.7 Performance Loss of Structurizing HPI

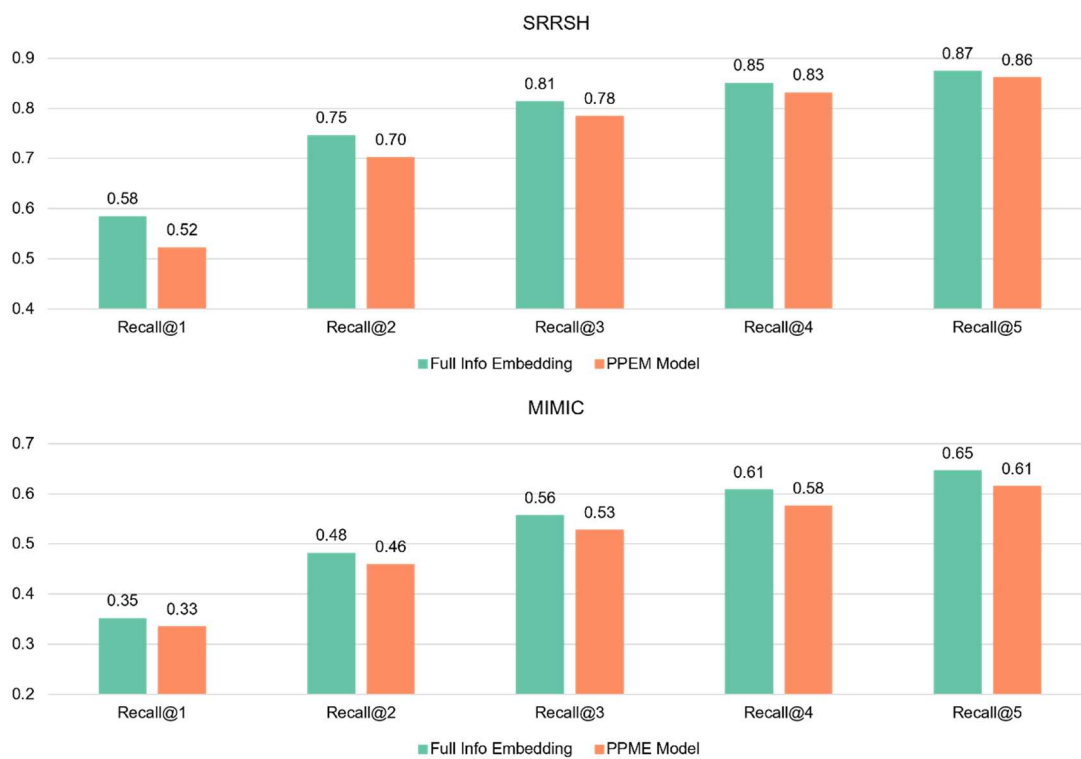

Supplementary Figure 6 Performance Loss of Structurizing HPI

In this study, we employed structured HPI to enumerate symptom information systematically. The completeness of our enumeration significantly impacts the performance of PPME LLM. In Supplementary Figure 6, "full info embedding" represents the disease prediction performance when directly using raw textual medical history information for embedding, combined with supervised learning methods. The "PPME Model" reflects prediction performance using structured HPI along with other textual medical historical data. Results show that the performance gap between the PPME Model and full info embedding remains around 5% in most cases. This observation demonstrates that our structured HPI comprehensively covers symptom information, and the structuring process does not lead to significant performance degradation.